\documentclass[11pt]{article}
\usepackage{eamt23}
\usepackage{times}
\usepackage{url}
\usepackage{latexsym}
\usepackage[small,bf]{caption} 
\usepackage{microtype}
\usepackage{graphicx}
\usepackage{subfigure}
\usepackage{ulem}
\usepackage{caption}
\usepackage{CJKutf8}
\usepackage{multirow}
\usepackage{algorithmic}
\usepackage{amsmath}
\usepackage{algorithm}
\usepackage{array}
\usepackage{booktabs}
\usepackage{float}
\usepackage{xcolor}
\definecolor{green}{RGB}{46,139,87}

\title{Learning Homographic Disambiguation Representation for Neural Machine Translation}

\newcommand*{\affaddr}[1]{#1} 
\newcommand*{\affmark}[1][*]{\textsuperscript{#1}}
\newcommand*{\email}[1]{\texttt{#1}}

\author{%
Weixuan Wang\affmark[1],Wei Peng\thanks{\quad Corresponding author} \affmark[1], Qun Liu\affmark[2] \\
\affaddr{\affmark[1]Artificial Intelligence Application Research Center, Huawei Technologies}\\
\email{\{wangweixuan2,peng.wei1\}@huawei.com}\\
\affaddr{\affmark[2]Noah's Ark Lab, Huawei Technologies}\\
\email{\{qun.liu\}@huawei.com}%
}

\date{}

\begin{document}
\maketitle
\begin{CJK}{UTF8}{gbsn}
\begin{abstract}
  Homographs, words with the same spelling but different meanings, remain challenging in Neural Machine Translation (NMT). While recent works leverage various word embedding approaches to differentiate word sense in NMT, they do not focus on the pivotal components in resolving ambiguities of homographs in NMT: the hidden states of an encoder. In this paper, we propose a novel approach to tackle homographic issues of NMT in the latent space. We first train an encoder (aka ``HDR-encoder'') to learn universal sentence representations in a natural language inference (NLI) task. We further fine-tune the encoder using homograph-based synset sentences from WordNet, enabling it to learn word-level homographic disambiguation representations (HDR). The pre-trained HDR-encoder is subsequently integrated with a transformer-based NMT in various schemes to improve translation accuracy. Experiments on four translation directions demonstrate the effectiveness of the proposed method in enhancing the performance of NMT systems in the BLEU scores (up to +2.3 compared to a solid baseline). The effects can be verified by other metrics (F1, precision, and recall) of translation accuracy in an additional disambiguation task. Visualization methods like heatmaps, T-SNE and translation examples are also utilized to demonstrate the effects of the proposed method.
\end{abstract}

\section{Introduction}


Homographs are words that have the same spelling but different meanings. For example, the word \textbf{\textit{``interest''}} is associated with a meaning expressing \textbf{\textit{``the feeling of wanting to know or learn about something or someone''}} (\textbf{\textit{``兴趣''}} in Chinese translation) and a sense of \textbf{\textit{``money paid regularly at a particular rate for the use of money lent''}} (\textbf{\textit{``利息''}} in Chinese translation). Despite recent advancements in Neural Machine Translation (NMT), resolving the ambiguity of homographs remains a research challenge in the face of data sparsity \cite{homo,analysis,encoder,jap,detect}. Handling ambiguity of homographs for NMT has been explored in \cite{homo}, \cite{weak}, \cite{improve} with a focus on word embeddings. In concord with \cite{encoder}, who pinpoint the pivotal role of encoder hidden states, we hypothesize that resolving the ambiguity of homographs should not be confined in the embedding space but need to investigate means to achieve this in the semantics-bearing latent space. 

The intuition here is that we may incorporate disambiguation knowledge into the latent space to handle homographs in NMT. In this paper, we propose a novel approach enabling an encoder (aka ``HDR-encoder'') to explicitly learn homographic disambiguation representation (HDR), which is further integrated with a transformer-based NMT model. The HDR-encoder is pre-trained using sentence-level semantic representation and homographic words coupled with related original-example sentence pairs from WordNet. In this way, homographic disambiguation knowledge can be learned by the latent space of the NMT. The pre-trained HDR-encoder is subsequently integrated with a transformer-based NMT in various ways to improve translation accuracy. 
Experimental results on four translation directions demonstrate consistency in the effectiveness of the proposed model in enhancing the BLEU score (up to +2.3 compared to a solid baseline). The effects can be further verified in the improvements of other scores (F1, precision, and recall) for the translation accuracy of homographs. Visualization methods like heatmaps, T-SNE and translation examples are utilized to demonstrate the effects of the proposed approach further.

In summary, our contribution are as follows:

\begin{itemize}
\item To our knowledge, we are the first to utilize the semantic representation of sentences in the latent space to handle the ambiguity of homographs in NMT. Experimental results show their effectiveness in improving both the BLEU scores and the other specific disambiguation scores;  
\item We develop a way to enable the proposed encoder to learn the word disambiguation representation of homographs with the help of synset example sentences extracted from WordNet, resulting in a further enhancement to the translation accuracy;    
\item We explore ways to incorporate the pre-trained homographic encoder with a standard transformer-based NMT model. The pre-trained HDR-encoder can be reused and enhanced by other researchers to resolve ambiguities in NMT.  
\end{itemize}

\section{Related Work}

Recent research has made significant progress in tackling the Word Sense Disambiguation (WSD) \cite{consec,wngc,sparsity} problem, which is to classify the sense of a word based on its contexts. WordNet \cite{wordnet}, the lexical database of reference, is widely designed as a data augmentation technique \cite{gloss,sensevocab} in the WSD task. Resolving ambiguous words is also essential for a machine translation (MT) task, as an MT model suffers from the issue associated with data sparsity and can not learn the correct sense of homographs. \cite{smt1}, \cite{smt2} and \cite{smt3} redesign the task of WSD for Statistical Machine Translation (SMT) and show that incorporating the predictions of a word sense disambiguation system into an SMT model can improve translation accuracy.


With NMT replacing SMT as the mainstream of MT systems, we have witnessed an increasing interest in the research community in resolving the ambiguity.  \cite{multi} propose an NMT model to solve the issue of ambiguity by taking a weighted sum of bidirectional RNN-generated sense-specific embeddings with the associated word embeddings. \cite{improve} propose integrating the lexical chains of the sense embeddings into NMT. \cite{weak} introduce three adaptive clustering algorithms for WSD to produce the sense vectors jointly used with word vectors for NMT. The word embedding of each token is concatenated with a vector representation of its sense obtained from one of the clustering methods to tackle the ambiguity problem. \cite{homo} provide empirical evidence that existing NMT can not correctly translate ambiguous words. They explore various methods to integrate the context of the input word with the word embeddings to differentiate the word sense before the encoding phase. \cite{korean} build a Korean WSD preprocessor base on a large-scale lexical semantic knowledge base without modifying the NMT model and conduct a series of Korean translation tasks to demonstrate validity. 
\cite{jap} use a pre-trained BERT to perform input word embedding before feeding them to the encoder of a Transformer to resolve the ambiguity problem in an English to Japanese translation task.



All approaches mentioned above focus on processing input word embeddings without using encoder hidden states. \cite{encoder} designed a classification task and discovered that encoder hidden states are more powerful representations than word embeddings when investigating the ability of NMT encoders and decoders to disambiguate word senses.  It is necessary to go beyond embedding space and investigate means in the semantics-bearing latent space for disambiguating homographs in NMT. 


\section{The Proposed Method}


The section proposes an encoder (``HDR-encoder'') to learn homographic disambiguation representation (HDR) knowledge, which is further integrated with a typical transformer-based NMT model in various ways.

\subsection{HDR-encoder}
\label{HDR}

The HDR-encoder is designed to learn disambiguation knowledge to distinguish the meanings of homographs in the latent space. The training of the HDR-encoder is divided into two steps. The first step is to train the HDR-encoder to learn the semantic representations of given sentences. We follow how \cite{infersent} propose to learn universal sentence representations from natural language inference (NLI) data. At the end of Step 1, the HDR-encoder is expected to output a sequence of hidden states of word tokens, capturing semantic features that differentiate sentences. In Step 2, we further fine-tune the HDR-encoder to learn fine-grained word-level disambiguation knowledge. 



\begin{figure}[!h]
    \centering
    \includegraphics[width=0.4\textwidth,height=0.2\textheight]{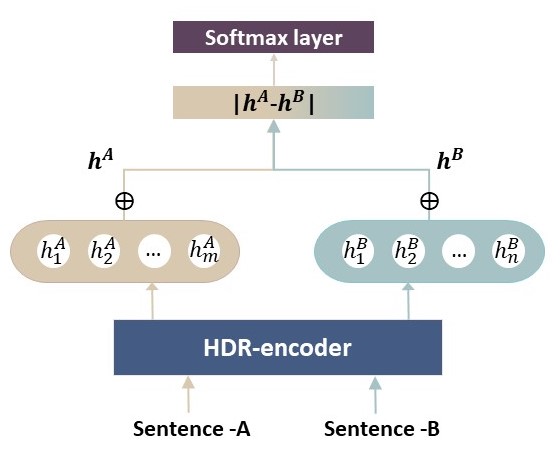}
    \caption{Training HDR-encoder on an NLI task enables it to differentiate sentences based on semantic representations. The output of the Softmax are labels of the NLI task during the training. }
    \label{nli}
\vspace{-0.5cm}
\end{figure}

\begin{figure*}[!h]
    \centering
    \includegraphics[width=0.9\textwidth,height=0.33\textheight]{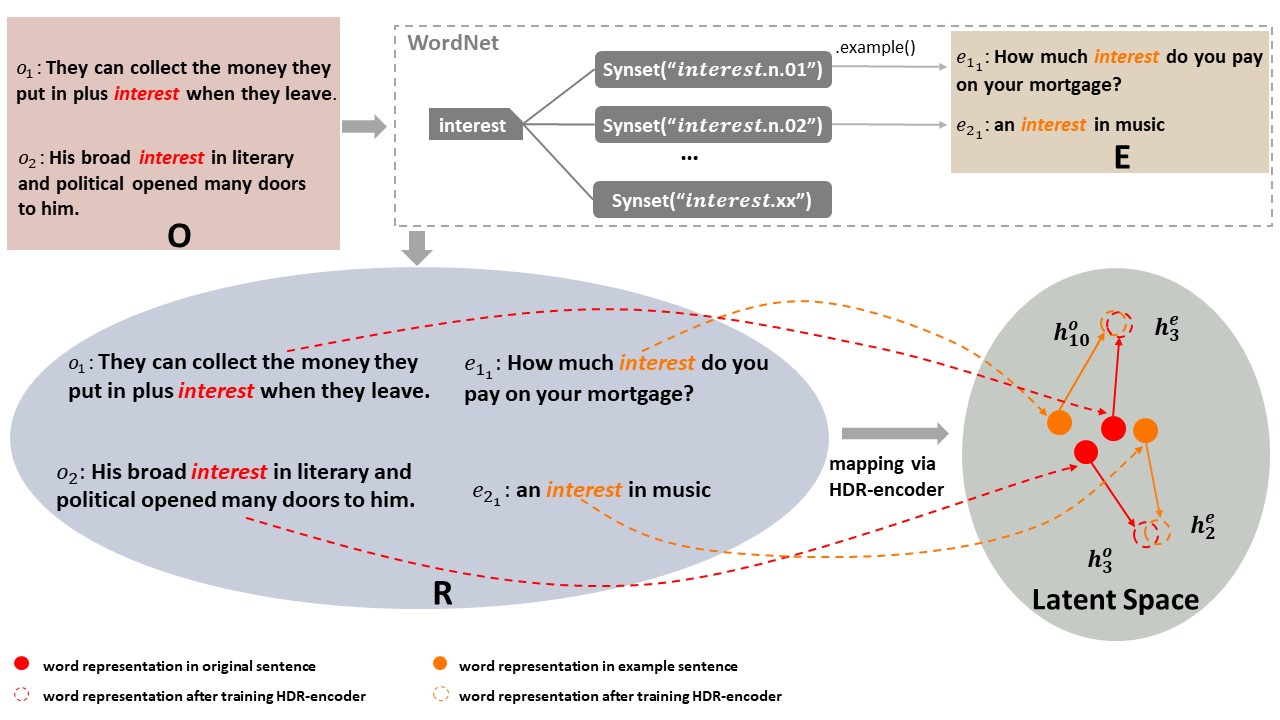}
    \caption{Train the HDR-encoder base on the WordNet at a fine-grained word level using the corpora constructed through WordNet. $h_{10}^o$ denotes the representation of the tenth word (the homograph ``interest'') in the original sentence $o_1$, and $h_3^e$ is the third word (the same homograph) in the example sentence $e_{1_1}$. The solid lines with an arrow between the solid circles in the latent space depict the process of minimizing their distances in the latent space.}
    \label{wsd}
\vspace{-0.6cm}
\end{figure*}

\subsubsection{Sentence Representation (SR) Learning}
\label{SRL}

Research works \cite{universal,infersent} show that a sentence can be represented as the sum of the hidden states of the consisting words. As shown in Figure~\ref{nli}, we train the HDR-encoder on an NLI task, enabling it to differentiate sentences based on semantics.


In this case, a shared encoder is used to produce the independent representations of two sentences. The sentence-A $\{x_1^{A}, x_2^{A}, ... , x_m^{A}\}$ and the sentence-B $\{x_1^{B}, x_2^{B}, ... , x_n^{B}\}$ are sent to the HDR-encoder at the same time. Their sentence-meaning representations in the latent space $\{h_1^{A}, h_2^{A}, ... , h_m^{A}\}$ and $\{h_1^{B}, h_2^{B}, ... , h_m^{B}\}$ are added to produce the sentence semantic representation $h^{A} = h_1^{A} + h_2^{A} + ... + h_m^{A}$, $h^{B} = h_1^{B} + h_2^{B} + ... + h_m^{B}$. Then the distance of two resulting semantic vectors are calculated by $|h^{A}-h^{B}|$, subsequently sent to the softmax layer in a typical NLI task. The HDR-encoder can learn sentence representations in this way.


\begin{algorithm}
\caption{Word Disambiguation Representation (WDR) Learning}
\label{algorithm-word}
\begin{algorithmic}[1]
\REQUIRE ~~\\
A set of training instances $O \big\{o_1,o_2,...,o_m\big\}$ with homographs labelled in WordNet synset IDs, HDR-encoder $H( \theta )$ with sentence-level representation, WordNet with synset examples $E_w \big\{e_{w_1},e_{w_2},...,e_{w_n}\big\}$
\ENSURE HDR-encoder with SR+WDR \\
\textbf{\underline{Data Preparation}}
\STATE \textbf{Initialization:} Disambiguation set $R\leftarrow\big\{ \big\}$
\STATE \quad \textbf{for all} $o_u \sim O$  \textbf{do}
\STATE \qquad \textbf{for each} homograph  $o_{u_w}$ in  all words 
\STATE \mbox{\qquad \quad $o_u\big\{o_{u_1},o_{u_2},...,o_{u_l}\big\}$ \textbf{do}}
\STATE \qquad \quad \textbf{for} {$v=1,2,...,n$} \textbf{do}
\STATE \qquad \qquad $R \leftarrow R \cup \big\{(o_u,e_{w_v})\big\}$  

\textbf{\underline{Word Level Training}}
\STATE \quad \textbf{for each} epoch $K = 1,2,...$ \textbf{do}
\STATE \qquad \textbf{for each} {pair $(o,e)$ in $R$} \textbf{do}
\STATE \qquad \quad Encode $(o,e)$ with $H( \theta ) \rightarrow   (h^o,h^e)$
\STATE \qquad \quad $\theta_K = \mathop{\arg\min}\limits_{R} cosdis(h_i^{o},h_j^{e}; \theta_{K-1} )$
\STATE \qquad \qquad \textbf{$where$} WDR for $o_i$ and $e_j$ are $h_i^o, h_j^e$  
\end{algorithmic}
\end{algorithm}

\subsubsection{Word Disambiguation Representation (WDR) Learning}


HDR-encoder is fine-tuned using original sentences coupled with example sentences extracted from WordNet \cite{wordnet} sharing identical homographs to learn word disambiguation representations (WDR). WordNet groups synonyms into synsets with different semantic meanings, providing a way to link homographs with related examples. As detailed in Algorithm~\ref{algorithm-word},  a set of original sentences $O \big\{o_1,o_2,...,o_m\big\}$  with homographs $(o_{u_w})$ labelled in terms of WordNet synset Ids is linked with sentence examples ($e_{w_v}$) with the same IDs. The coupled sentence pairs $(o_u,e_{w_v})$ are encoded to produce disambiguation representation in the form of $(h^o,h^e)$. The encoder is trained using a loss function depicted in Equation~\ref{eq1} by minimizing the cosine distance between word-level DR $h_i^{o}$ and $h_j^{e}$, with the $i$ and $j$ representing the position of homograph in both sentences. 

\vspace{-0.4cm}
\begin{equation}
\theta_K = \mathop{\arg\min}\limits_{R} cosdis(h_i^{o},h_j^{e}; \theta_{K-1} )
\label{eq1}
\end{equation}
\vspace{-0.4cm}

A detailed example is illustrated in Figure~\ref{wsd}. The homograph \textit{\textbf{``interest''}} in the original sentence \textit{``They can collect the money they put in plus \textbf{interest} when they leave.''} corresponds to a linked example sentence \textit{``How much \textbf{interest} do you pay on your mortgage?''} in the WordNet via the shared synset ID (``interest.n.01''). By minimizing the distance between the word level DR of the homograph (i.e., \textit{\textbf{``interest''}}) in the latent space, homographs with the same meaning can be clustered together. In this way, an HDR-encoder is expected to resolve the ambiguity of homographs more effectively than a traditional encoder in the vanilla transformer.

\subsection{Integration of HDR-encoder into Neural Machine Translation}

\begin{figure*}[h]
\centering                                                          
\subfigure[HDR-NMT\_add]{\label{add}\includegraphics[scale=0.25]{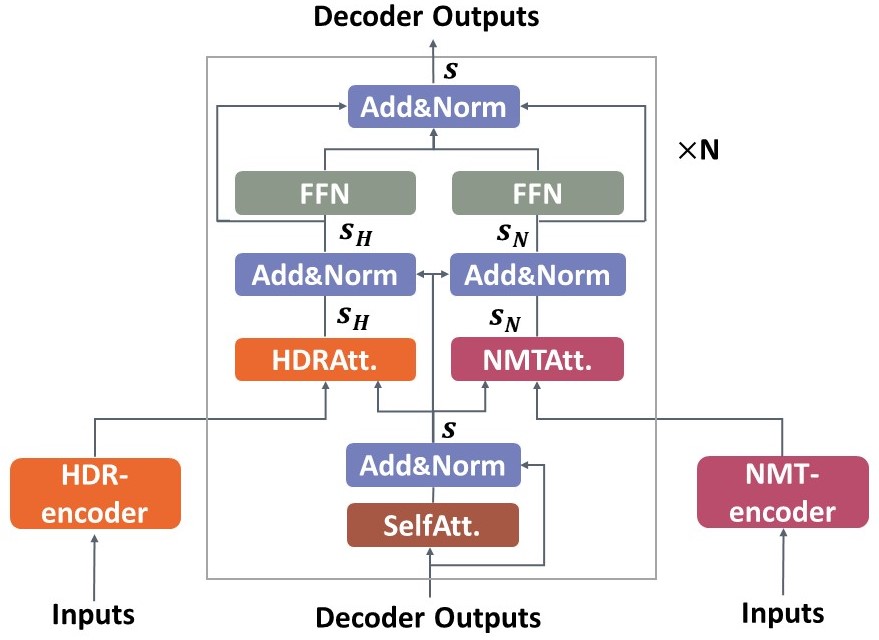}}               
\subfigure[HDR-NMT\_gate]{\label{gate}\includegraphics[scale=0.25]{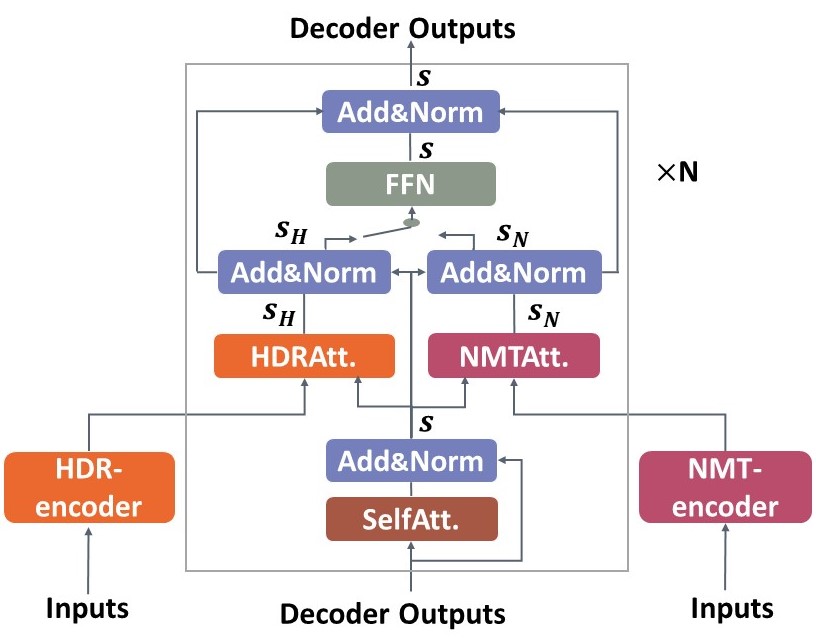}}
\subfigure[HDR-NMT\_cascade]{\label{cascade}\includegraphics[scale=0.25]{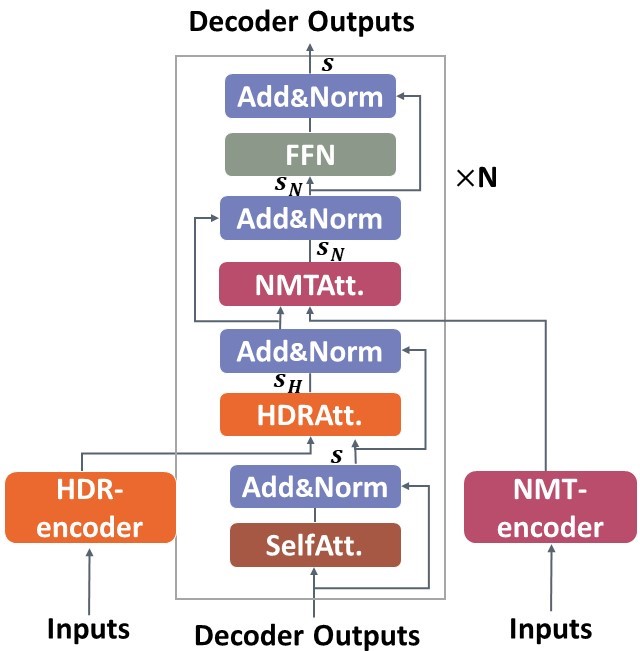}} 
\caption{The multiple ways to integrate the HDR-encoder with NMT. The ``NMT-Encoder'' denotes the encoder of NMT, the ``SelfAtt.'' denotes the self-attention of the decoder, the ``HDRAtt.'' denotes the cross-attention between the HDR-encoder and the decoder, the ``NMTAtt.'' denotes the cross-attention between the encoder of NMT and the decoder, and the ``FFN'' denotes the feed-forward layer. (a) shows the architecture of the HDR-NMT\_add. (b) is the architecture of the HDR-NMT\_gate. (c) is the architecture of HDR-NMT\_cascade.}  
\label{NMT} 
\vspace{-0.5cm}
\end{figure*}


We propose several methods to incorporate the HDR-encoder into a typical NMT: \textit{HDR-NMT\_add}, \textit{HDR-NMT\_gate} and \textit{HDR-NMT\_cascade} (Figure~\ref{NMT}). The specific mechanism is detailed as below.





\subsubsection{\textit{HDR-NMT\_add}}

As illustrated in Figure~\ref{add}, the inputs are fed into the HDR-encoder and an NMT encoder separately to calculate cross attention (in ``HDRAtt.'' and ``NMTAtt.'') with their results $s_{H}$ and $s_{N}$ added to produce the output $s$. The layer norm (aka ``LN'') and the feed-forward network (``FFN'') are processed separately, as depicted in Equations~\ref{norm-homo}-\ref{norm-nmt}, where $s$ is the input of cross attention on the side of the NMT decoder. Then the outputs of cross attention $s_H$, $s_N$ are added to produce the final output (Equation~\ref{homo-nmt}) sent to the next layer for further calculation.  
\vspace{-0.2cm}
\begin{equation} 
s_H = LN(s + s_H)
\label{norm-homo}
\end{equation}
\vspace{-0.7cm}
\begin{equation}
s_N = LN(s + s_N)
\label{norm-nmt}
\end{equation}
\vspace{-0.7cm}
\begin{equation}
s\!=\!LN(FFN(s_H)\!+\!FFN(s_N)\!+s_H\!+\!s_N)
\label{homo-nmt}
\vspace{-0.2cm}
\end{equation}

\subsubsection{\textit{HDR-NMT\_gate}}
\label{HDR-NMT-gate}
Inspired by the effectiveness of HDR-NMT\_add schema, we design a gating mechanism to explore the proportional contributions of HDR-encoder and NMT-encoder. As illustrated in Figure~\ref{gate}, the inputs are fed into the HDR-encoder and an NMT encoder separately to calculate cross attention (in ``HDRAtt.'' and ``NMTAtt.'') with their results $s_{H}$ and $s_{N}$ gated to produce the output $s$. 

\vspace{-0.4cm}
\begin{equation}
s = Gate(s_H + s_N)
\end{equation}
\vspace{-0.6cm}
\begin{equation}
s\!=\!LN(FFN(s)\!+\!s_N\!+s_H)
\end{equation}
\vspace{-0.7cm}

The gate mechanism can be implemented in various ways, for example, using a fixed value or an activation function like sigmoid. It is discovered from the experiments that a gate mechanism with an empirical value of 0.5 produces results outperforming those from a sigmoid function.  


\subsubsection{\textit{HDR-NMT\_cascade}} 

Apart from a parallel integration of ``HDRAtt.'' and ``NMTAtt.'' as mentioned above, we also explore their sequential alignments. As illustrated in Figure~\ref{cascade}, we prepend the HDR-encoder to the NMT-encoder considering the task aspects of NMT. The cross attention of the HDR-encoder ($s_H$) is calculated, followed by the one for NMT-encoder ($s_N$) (in Figure~\ref{cascade}). The final normalization is depicted in Equation~\ref{cas}:

\vspace{-0.4cm}
\begin{equation} 
s = LN(FFN(s_N)+s_N)
\label{cas}
\end{equation}
\vspace{-1cm}


\section{Experiments}

The HDR-encoder is trained using the SNLI dataset \footnote{https://nlp.stanford.edu/projects/snli/} and the MultiNLI dataset \footnote{https://cims.nyu.edu/~sbowman/multinli/} with a total size of 0.9M to learn sentence level representation (Section~\ref{SRL}). We select the annotated dataset SemCor \footnote{https://www.gabormelli.com/RKB/SemCor\_Corpus}, which contains 352 sentences, as seed data to construct the set $R$ to enable the HDR-encoder to learn the word-level disambiguation representations (depicted as the data preparation in Algorithm~\ref{algorithm-word}). The original SemCor data is extended to 0.1M, accommodating homograph-loaded synset examples from WordNet.

We further train and evaluate the proposed approach on four language directions, namely, English (EN) $ \rightarrow $ Russian (RU), English (EN) $ \rightarrow $ Chinese (ZH), English (EN) $ \rightarrow $ German (DE) and English (EN) $ \rightarrow $ French FR), on a range of popular datasets. For the EN $ \rightarrow $ DE and EN $ \rightarrow $ FR  task, we select WMT14 data as the training and the development set with newstest2014 as the test set. A standard data cleaning method is applied to clean the training data, within which we randomly sample some sentences as the development (dev.) set. 
Following the same fashion, the WMT17 (with newstest2017) and WMT19 (with newstest2019) are selected for EN $ \rightarrow $ RU and  EN $ \rightarrow $ ZH, respectively. The experiments for the HDR-encoder and NMT use the same vocabulary, which is not shared for the source and target language for each language direction. The statistics of the datasets are illustrated in Table~\ref{dataset} at Appendix \ref{detail}.

\begin{table*}[]\small
\centering
\begin{tabular}{lccccc}
\hline
\textbf{Model} & \textbf{EN $ \rightarrow $ RU} & \textbf{EN $ \rightarrow $ ZH} & \textbf{EN $ \rightarrow $ DE} & \textbf{EN $ \rightarrow $ FR} & \textbf{AVG-Increase} \\ \hline
BiLSTM  & -  & - & 21.80  & 32.39 & \\
K-means+Att   & - & -  & 23.85 & - &  \\
Fairseq  & 23.9 & 25.5  & 26  & 38.3 &  \\
HDR-NMT\_add  & 25.1(+1.2) & 26.2(+0.7)  & \textbf{26.9(+0.9)}  & \textbf{39.0(+0.7)} & +0.88 \\
HDR-NMT\_gate  & \textbf{26.2(+2.3)}  & \textbf{26.6(+1.1)} & 26.7(+0.7) & 38.8(+0.5) & \textbf{+1.15} \\ 
HDR-NMT\_cascade  & 25.5(+1.6) & 26.5(+1.0) & 26.5(+0.5) & 38.4(+0.1) & +0.80 \\ 
\hline
\end{tabular}
\caption{The overall experimental results of the proposed approach compared with other models, evaluated on newstest17-EN $ \rightarrow $ RU, newstest19-EN $ \rightarrow $ ZH, newstest14-EN $ \rightarrow $ DE and newstest14-EN $ \rightarrow $ FR. The numbers in parentheses represent the improvements of BLEU scores over the baseline Fairseq BLEU score.}
\label{overall}
\vspace{-0.4cm}
\end{table*}

The architecture of the HDR-encoder and the NMT-encoder follows the one proposed in the vanilla Transformer \cite{transformer}. The training of the HDR-encoder for each language direction is on 1 NVIDIA V100 GPU. Experiments on the HDR-NMT tasks are based on servers with 4 NVIDIA V100 GPUs, following \cite{transformer} in hyper-parameter configuration settings. The number of parameters are illustrated in Table~\ref{para} at Appendix \ref{number}. The training of the HDR-NMT is as fast as baselines as the parameters of the HDR-encoder are frozen and do not participate in the back propagation. It took approximately 25 hours to train a baseline EN $ \rightarrow $ ZH NMT model, while training HDR-NMT requires 24-29 hours (all in 4 NVIDIA V100 GPUs). The pre-training of the HDR-encoder takes around 1 hour.

Early stopping is applied to all training. The maximum batch size is set to 6044, and the dropout probability is set to 0.1. We use the sacreBLEU \footnote{BLEU+case.mixed+numrefs.3+smooth.exp+tok.13a+ version.1.4.12.} to measure the accuracy of the HDR-NMT. We compare the proposed approach against the following baselines: \textbf{Fairseq} (baseline) is a sequence modeling toolkit \cite{fairseq}. \textbf{BiLSTM} model the context-aware word embedding to differentiate the word sense in \cite{homo}. \textbf{K-means+Att} propose using adaptive clustering algorithms for WSD in the NMT system \cite{weak}.    


\begin{table}[]\footnotesize
\centering
\setlength\tabcolsep{3pt}{
\begin{tabular}{lccc}
\hline
\textbf{Model} & \textbf{add} & \textbf{gate} & \textbf{cascade} \\ \hline
Fairseq  & 23.9  & 23.9  & 23.9   \\ 
SR  & 24.1(+0.2)  & \textbf{25.0(+1.1)}  & \textbf{24.8(+0.9)}  \\
SR+WDR & 24.2(+0.3)  & 25.6(+1.7) & \textbf{25.5(+1.6)} \\ 
FT-SR  & 24.9(+1.0)  & \textbf{25.6(+1.7)} & 25.3(+1.4)  \\ 
FT-SR+WDR  & 25.1(+1.2)  & \textbf{26.2(+2.3)} & 25.5(+1.6) \\ \hline
\end{tabular}}
\caption{The specific results of EN $ \rightarrow $ RU translation task. The ``SR'' represents the HDR-encoder's experiment with only learned sentence-level representations. ``SR+WDR'' represents the study of the HDR-encoder with learned representations at both sentence and word levels. ``FT-SR'' and ``FT-SR+WDR'' represent the NMT models fine-tuned from the baseline with HDR-encoder on different levels of representations.   }
\label{en2ru}
\end{table}

\begin{table}[h]\footnotesize
\centering
\setlength\tabcolsep{3pt}{
\begin{tabular}{lc}
\hline
\textbf{Model} & \textbf{EN $ \rightarrow $ ZH}   \\ \hline
Fairseq (12-layer encoder)  & 25.1    \\ 
HDR-NMT (12-layer encoder) & 26.4 (+1.3)  \\  \hline
\end{tabular}}
\caption{Results of using a pre-trained language model to implement the HDR-encoder. }
\label{roberta}
\vspace{-0.6cm}
\end{table}

\vspace{-0.1cm}
\subsection{Experimental Results}
The overall results are shown in Table~\ref{overall}. Our HDR-NMT method achieves consistent results, significantly outperforming the Fairseq baseline by +2.3, +1.1, +0.9 and +0.7 BLEU points respectively in the involved four translation tasks (EN $ \rightarrow $ RU, EN $ \rightarrow $ ZH , EN $ \rightarrow $ DE and EN $ \rightarrow $ FR). Compared to other disambiguation methods (i.e., BiLSTM and K-means+Att), our method achieves +5.1 and +3.05 BLEU scores higher in the EN $ \rightarrow $ DE task. 

The HDR-NMT\_gate outperforms other integration schemes in average BLEU enhancement (AVG-Increase in Table~\ref{overall}) in the four language directions. It may ascribe to the design of the HDR-NMT\_{gate}, in which the FFN is designed to take a combined output from a gated configuration of the HDR-encoder and the NMT-encoder instead of dealing with $s_{H}$ or $s_{N}$ independently. Explaining the rationale behind the observation is out of the scope of this work, requiring a future study. 


\begin{table*}[h]\footnotesize
\centering
\begin{tabular}{c|c|ccc|ccc}
\hline
\multicolumn{1}{c|}{\multirow{2}{*}{\textbf{Language}}} & \multicolumn{1}{c|}{\multirow{2}{*}{\textbf{Model}}} & \multicolumn{3}{c|}{\textbf{Homograph}} & \multicolumn{3}{c}{\textbf{All-word}} \\ \cline{3-8}
\multicolumn{1}{c|}{}   & \multicolumn{1}{c|}{} & \textbf{F1}     & \textbf{Precision}  & \textbf{Recall}  & \textbf{F1}     & \textbf{Precision}  & \textbf{Recall} \\ \hline
\multirow{3}{*}{EN $ \rightarrow $ ZH} & baseline             & 45.7  & 45.0       & 46.4   & 51.5  & 53.2      & 49.8  \\
                       & SR        & 46.5(+0.8)  & 46.2(+1.2)      & 46.8(+0.4)   & 52.4(+0.9)  & 54.3(+1.1)      & 50.7(+0.9)  \\
                       & SR+WDR            & 47.5(+1.8)  & 46.5(+1.5)      & 48.5(+2.1)   & 52.6(+1.1)  & 54.4(+1.2)      & 51.0(+1.2)   \\ \hline
\multirow{3}{*}{EN $ \rightarrow $ RU} & baseline             & 39.8  & 40.8      & 38.8   & 44.8  & 46.5      & 43.2  \\
                       & SR       & 40.6(+0.8)  & 41.7(+0.9)      & 39.8(+1.0)   & 46.1(+1.3)  & 47.9(+1.4)      & 44.4(+1.2)  \\
                       & SR+WDR            & 41.3(+1.5)  & 42.1(+1.3)      & 40.6(+1.8)   & 46.3(+1.5)  & 48.1(+1.6)      & 44.6(+1.4)  \\ \hline
\multirow{3}{*}{EN $ \rightarrow $ DE} & baseline             & 43.4  & 45.2      & 41.8   & 52.3  & 54.6      & 50.1  \\
                       & SR        & 44.4(+1.0)  & 46.3(+1.1)      & 42.8(+1.0)   & 52.8(+0.5)  & 55.0(+0.4)       & 50.8(+0.7)  \\
                       & SR+WDR            & 45.1(+1.7)  & 46.8(+1.6)      & 43.4(+1.6)   & 53.0(+0.7)   & 55.2(+0.6)      & 50.9(+0.8)  \\ \hline
\multirow{3}{*}{EN $ \rightarrow $ FR} & baseline             & 38.9  & 39.6      & 38.2   & 51.4  & 52.6      & 50.3  \\
                       & SR        & 41.5(+2.6)  & 42.2(+2.6)      & 40.9(+2.7)   & 51.7(+0.3)  & 52.8(+0.2)      & 50.6(+0.3)  \\
                       & SR+WDR            & 41.7(+2.8)  & 42.4(+2.8)      & 41.0(+2.8)    & 51.8(+0.4)  & 53.0(+0.4)       & 50.7(+0.4) \\ \hline
\end{tabular}
\caption{Experimental results for disambiguation of homographs and all words in terms of F1, precision, and recall with p < 0.05, indicating statistical significance across all metrics. }
\label{F1}
\vspace{-0.4cm}
\end{table*}

\vspace{-0.1cm}
\subsection{Effects of SR and WDR Learning}
To further investigate the effects of sentence representation (SR) and word disambiguation representation (WDR) learning depicted in Section~\ref{HDR}, we conduct an ablation study for the translation task with a language direction from EN $ \rightarrow $ RU. We train the proposed HDR-NMT schemes with HDR-encoder to learn SR alone and in combination with WDR (``SR+WDR'') and compare them with the results obtained from the baseline. Table~\ref{en2ru} shows that learning WDR further contributes +0.7 BLEU scores (24.8 to 25.5) to the HDR-NMT\_{cascade} on the top of its enhancement of +0.9 from a baseline (23.9 to 24.8). When learning SR alone, a significant gain of +1.1 (from 23.9 to 25.0) comes from the HDR-NMT\_{gate}. 

Finetuning the HDR-NMT with the best checkpoint of the baseline (``FT-SR'' and ``FT-SR+WDR'') produces the most significant results in HDR-NMT\_{gate}, achieving +1.7 (23.9 to 25.6)  and +2.3 (23.9 to 26.2) over the baseline respectively. Similar results are recorded in Tables ~\ref{en2zh}-\ref{en2de} of Appendix \ref{effect} for EN $ \rightarrow $ ZH and EN $ \rightarrow $ DE translation tasks. SR and WDR learning achieve consistent results, indicating the effectiveness of the proposed approach.                   

\begin{figure*}[h]\small
\centering                                                          
\subfigure[baseline-encoder]{\label{baseline-encoder}\includegraphics[scale=0.38]{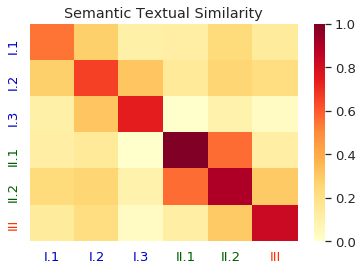}}    
\subfigure[NMT-encoder]{\label{NMT-encoder}\includegraphics[scale=0.38]{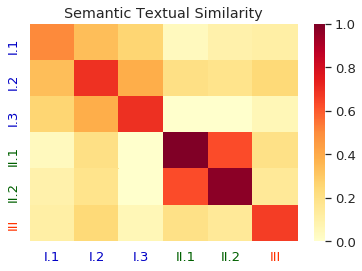}}
\subfigure[HDR-encoder]{\label{HDR-encoder}\includegraphics[scale=0.38]{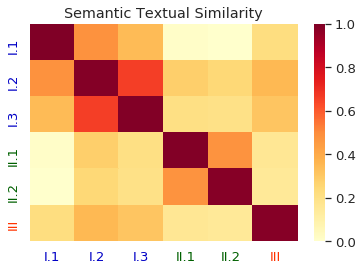}} 
\caption{Visualizing the capability of the proposed method in grouping a homograph (``right'') with different meanings. (a) The heatmap result of the baseline encoder. (b) The heatmap result of the fine-tuned NMT-encoder. (c) The heatmap result of the HDR-encoder. \color{blue}I.1 \color{black}: But the government had made the \color{blue}right \color{black} call. \color{blue}I.2\color{black}: But which is the \color{blue}right \color{black}template? \color{blue}I.3\color{black}:What is the \color{blue}right \color{black} response to a collapse? \color{green}II.1\color{black}: The mirror, viewed from the \color{green}right\color{black}, reflected a gold colour. \color{green}II.2\color{black}: Now look at the photo on the \color{green}right\color{black}. \color{red}III\color{black}: You should take the \color{red}right \color{black}fork.}
\label{right}   
\vspace{-0.4cm}
\end{figure*}

\begin{figure}[h]
\centering                                                          
\subfigure[baseline-encoder]{\label{baseline-tsne}\includegraphics[scale=0.28]{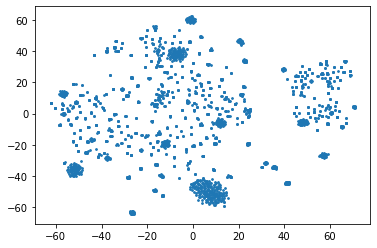}}
\subfigure[HDR-encoder]{\label{HDR-tsne}\includegraphics[scale=0.28]{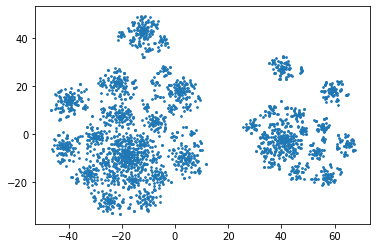}} 
\caption{T-SNE visualization of the capability of the proposed method in clustering the homograph with different synsets in the latent space. (a) the outputs of the baseline encoder for the sampled data containing the homograph (``right''). (b) the outputs of the HDR-encoder for the sampled data containing the homograph (``right''), showing an apparent clustering effect.}
\label{t-sne}     
\vspace{-0.4cm}
\end{figure}

\begin{table*}[h]\footnotesize
\begin{tabular}{p{130pt}p{95pt}p{95pt}p{95pt}}
\hline
\specialrule{0em}{1pt}{1pt}
\textbf{Source}  & \textbf{Baseline Output}  & \textbf{HDR-NMT Output}  & \textbf{Ground Truth} \\ \hline
\specialrule{0em}{1pt}{1pt}
Natural disasters make \color{red}\textbf{logistics} \color{black} even more complicated. & 自然灾害使\color{red}\textbf{后勤}\color{black}变得更加复杂。 & 自然灾害使\color{red}\textbf{物流}\color{black}更加复杂。 & 自然灾害致使\color{red}\textbf{物流}\color{black}变得更加艰难。 \\ \hline
\specialrule{0em}{1pt}{1pt}
\color{red}\textbf{Interest } \color{black} remained high after the hearing.   & 在听审之后,\color{red}\textbf{利息}\color{black}仍然很高。& 听证会之后,\color{red}\textbf{兴趣}\color{black}仍然很高。 & 听证会结束后，\color{red}\textbf{关注度}\color{black}仍居高不下。 \\ \hline
\specialrule{0em}{1pt}{1pt}
Bodies of some victims were found trapped under the \color{red}\textbf{rubble} \color{black} of collapsed buildings & 一些受害者的尸体被困在倒塌建筑的\color{red}\textbf{瓦砾}\color{black}之下 & 发现一些受害者的尸体被困在倒塌的建筑物的\color{red}\textbf{废墟}\color{black}下 & 一些遇难者的尸体被压在坍塌建筑的\color{red}\textbf{废墟}\color{black}下   \\ \hline
\specialrule{0em}{1pt}{1pt}
Pompeo has visited North Korea three times already this year,but his \color{red}\textbf{last} \color{black} trip did not go well.  & 蓬佩奥今年已经三次访问朝鲜,但他的\color{red}\textbf{上一次}\color{black}访问并不顺利。 & 蓬佩奥今年已经三次访问朝鲜,但他的\color{red}\textbf{最后一次}\color{black}访问并不顺利。& 今年，蓬佩奥已经三次访问朝鲜，但是他的\color{red}\textbf{最后一次}\color{black}行程并不顺利。\\ \hline
\specialrule{0em}{1pt}{1pt}
It is intolerant of those who \color{red}\textbf{speak their minds} \color{black}  & 不容忍那些\color{red}\textbf{表达自己心态}\color{black}的人。 & 这是不能容忍那些\color{red}\textbf{说自己心声}\color{black}的人。 & 这里容不得\color{red}\textbf{直言}\color{black}的人。 \\ \hline
\specialrule{0em}{1pt}{1pt}
It is understood they will feature a powerful cannon, an array of anti-aircraft and anti-ship missiles as well as some stealth technologies, such as reduced radar, infrared and acoustic \color{red}\textbf{signature} \color{black}. & 了解到它们将使用强力的大炮、一系列防空和反舰导弹以及一些隐形技术,例如减少雷达、红外和声学\color{red}\textbf{签名}\color{black}。 & 据了解,它们将包括强大的火炮、一系列防空导弹和反舰导弹以及一些隐形技术,如减少雷达、红外线信号和声学\color{red}\textbf{信号}\color{black}。 & 据了解，这艘护卫舰将拥有强大的火力、大批防空和反舰艇导弹以及一些隐形技术、如减少雷达、红外线和声学\color{red}\textbf{信号}\color{black}。 \\ \hline
\specialrule{0em}{1pt}{1pt}
They use a sophisticated echo-location technique to pinpoint \color{red}\textbf{bugs} \color{black} and obstacles in their flight path.   & 他们使用精密的回呼定位技术,在其飞行路径中查明\color{red}\textbf{窃听}\color{black}和障碍物。 & 他们使用精密的回声位置技术,在其飞行道路上发现\color{red}\textbf{虫子}\color{black}和障碍。 & 它们能够利用复杂的回声定位技术准确定位飞行路径中的\color{red}\textbf{小虫}\color{black}和障碍物。 \\ \hline
\specialrule{0em}{1pt}{1pt}
But the UK Government has also been stepping up its preparations for a possible \color{red}\textbf{no-deal} \color{black} scenario.  & 但是,联合王国政府也一直在加紧为可能的\color{red}\textbf{不作交易}\color{black}的设想作准备。 & 但联合王国政府也一直在加紧准备,以备可能出现的\color{red}\textbf{不达成协议}\color{black}的情况。  & 但是英国政府也在加紧为可能出现的\color{red}\textbf{无协议}\color{black}局面做准备。\\ \hline       
\end{tabular}
\caption{The comparable examples extracted from the outcomes of the newstest19 EN $ \rightarrow $ ZH.}
\label{example}
\vspace{-0.4cm}
\end{table*}


\vspace{-0.1cm}
\subsection{Pre-trained Language Model as HDR-Encoder}
\label{PLM}
An additional study is performed to investigate the effect of implementing the HDR-encoder using a pre-trained language model. We initiate the HDR-encoder using Roberta, which consists of a 12-layer encoder \cite{roberta}. The vocabulary of Roberta is shared in the baseline and the HDR-NMT, both with a 12-layer encoder. Table~\ref{roberta} demonstrates a positive effect in the EN $ \rightarrow $ ZH translation task where the HDR-encoder of HDR-NMT is finetuned on the 12-layer transformer encoder.

\vspace{-0.1cm}
\subsection{Additional NMT Encoder Does Not Work}
It is interesting to know if adding an additional NMT encoder could achieve a similar effect to that of the proposed HDR-encoder. We design a comparison study to demonstrate the effect of dual NMT encoders in HDR-NMT\_{add} for the EN $ \rightarrow $ ZH task. The negative results shown in Table~\ref{dual} of Appendix \ref{additional} indicate that simply adding NMT-encoder does not work.   


\vspace{-0.1cm}
\subsection{Experimental Results in Other Metrics}
 
To further verify the effect of the proposed method in resolving homographic ambiguity, we compare \textbf{HDR-NMT} with the baseline on a list of homographs extracted from Wikipedia \footnote{https://en.wikipedia.org/wiki/List\_of\_English\_homographs} on the four language directions. We follow the method in \cite{homo} to evaluate the effects using other metrics, including F1, precision, and recall, as shown in Table~\ref{F1}. The improvement of our proposed method over the baseline model is consistently significant for HDR-NMT with learned SR alone and in combination with WDR (``SR+WDR'').     


\vspace{-0.1cm}
\subsection{Human Evaluation}
Human evaluation is also conducted to illustrate the effectiveness of the proposed method. One hundred sentences are sampled randomly and scored by linguistics experts using a score scale from 1 to 10 based on translation quality and fluency. As shown in Table~\ref{human} (Appendix), HDR-NMT outperforms the baseline in translation quality by a wide margin (Reference 8.075 vs. HDR-NMT 7.700 vs. Baseline 6.050).

\vspace{-0.1cm}
\subsection{Visualization}

Figure~\ref{right} visualizes the capability of the proposed method in grouping a homograph with different meanings in the hidden space using heatmaps. The strength of the shades in the heatmaps represents the degree of semantic similarity for a homograph (\textbf{``right''} in this case) with different meanings embedded in a sentence. We sample six sentences from the training dataset with the \textbf{``right''} in the first three sentences marked in blue, indicating the meaning of \textbf{``correct (正确的 in Chinese)''}. The same homograph \textbf{``right''} in the last two sentences colored in green means \textbf{``the right-hand (右侧 in Chinese)''}. Please note that although the homograph \textbf{``right''} in the last sentence (III) colored in red means ``right-hand direction'' of a road junction in the reference data, it can be used to indicate a ``correct'' fork as an item of cutlery. It is clearly shown that the proposed HDR-NMT with an HDR-encoder (Figure~\ref{HDR-encoder}) and an NMT-encoder (Figure~\ref{NMT-encoder}) achieve higher semantic similarities than those from the baseline (Figure~\ref{baseline-encoder}).

We sample 3,000 sentences containing the homograph ( ``right'' ) and visually compare the distributions of the homographic word representations in the latent space using T-SNE, as shown in Figure \ref{t-sne}. Figure~\ref{baseline-tsne} illustrates the outputs of the baseline encoder for the sampled data without any HDR learning. The output is primarily a random distribution with some degrees of regional aggregations. In comparison, a clear agglomerative pattern can be observed for the outputs of the HDR-encoder (Figure~\ref{HDR-tsne}). There are 22 synsets of the homograph ``right'' in the training data set ``SemCor''. Although we do not know how many meanings are presented in this batch of sample data, the proposed HDR learning can produce clustered representations in the hidden space, resulting in ambiguity resolution in NMT.  

Table~\ref{example} captures the examples from the outputs of the proposed approach in comparison with the baseline for the EN $ \rightarrow $ ZH task. It can be observed that the HDR-NMT can distinguish the meanings of homographs, including ``interest'', ``rubble'' and ``logistics'', etc. For example, the homograph ``bug'' is accurately translated into the meaning of ``insect'' (小虫 in Chinese), which the baseline ill-translates into ``窃听'' (``eavesdropping'') in comparison. Please refer to the heatmaps and more examples shown in Figure~\ref{logistics} and Table~\ref{example2} in Appendix \ref{vis} for a more detailed visualization.

\section{Conclusion}


In this paper, we propose a novel approach to resolve the problem of homographs in the latent space by leveraging the knowledge learned by an HDR-encoder, which can be used for disambiguation. The pre-trained HDR-encoder is subsequently integrated with a transformer-based NMT to improve the translation accuracy. Experimental results demonstrate that the proposed HDR-NMT can significantly improve translation accuracy by reducing ambiguous translation errors and boosting the BLEU scores on the EN $ \rightarrow $ RU, EN $ \rightarrow $ ZH, EN $ \rightarrow $ DE, and EN $ \rightarrow $ FR translation tasks. The effects of the proposed approach can further be verified in the improvements of other disambiguation metrics, including F1, precision, and recall in an additional task. Visualization techniques including heatmaps, T-SNE and translation examples are used to further demonstrate the proposed method's effects.     

\section{Limitations}
Although the proposed method effectively resolves the ambiguity of homographs in the latent space for neural machine translation, we do not investigate the potential advantages of word sense embeddings. We plan to probe into the effects of jointly training word sense embeddings in conjunction with HDR. Furthermore, the proposed method is constrained by several factors: Firstly, the gating mechanism mentioned in Section~\ref{HDR-NMT-gate} may be further improved by exploring an optimal configuration of contributions from the HDR-encoder and the NMT-encoder. In addition, it is noted that the level of improvement can vary for different language directions. We have yet to establish the rationale behind this observation, and we conjecture this may relate to the number of homographs in various translation tasks. As our ultimate goal is to extend the proposed method to a multilingual setting, we should further enhance the representation learning of SR and WDR with more NLI and homograph-labeled multilingual data. Moreover, the experiment in Section~\ref{PLM} demonstrates the positive effect of implementing the HDR-encoder using a pre-trained language model on a translation direction (EN $ \rightarrow $ ZH), but how to scale this up warrants further investigation.




\bibliography{eamt23}
\bibliographystyle{eamt23}

\clearpage

\appendix

\label{sec:appendix}

\section{Appendix}

\subsection{\textit{HDR-NMT\_selection}}

\begin{figure}[!h]
    \centering
    \includegraphics[width=0.35\textwidth,height=0.17\textheight]{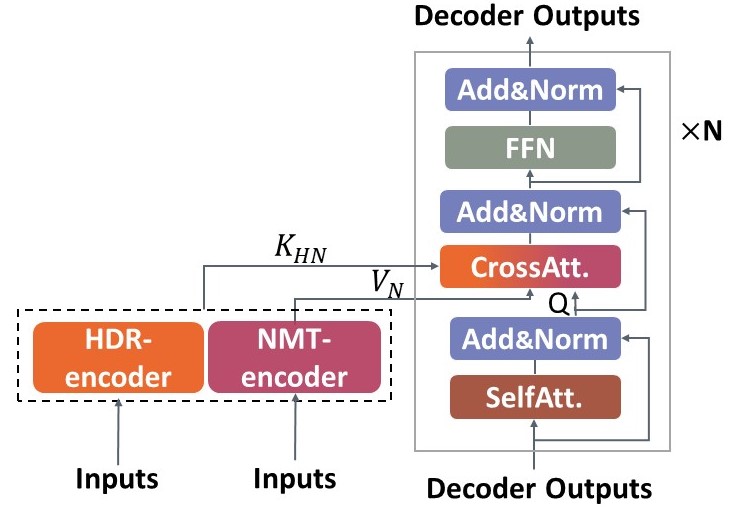}
    \caption{The architecture of HDR-NMT\_selection. }
    \label{selection}
\end{figure}
In this section, we introduce a new design (aka HDR-NMT\_selection) of how to integrate the HDR-encoder with the NMT. We unify the calculation of the cross attention of the HDR-encoder and the NMT-encoder in an attempt to reduce computation overhead by modifying key $K_{HN}$ as the concatenation of the outputs of the HDR-encoder with the outputs of the NMT encoder. Moreover, the value $V_N$ is still the output of the NMT encoder. We use the disambiguation knowledge from the HDR encoder to select the appropriate translation information from the NMT encoder. \textit{HDR-NMT\_selection} only has one calculation process of the cross attention shown in Figure~\ref{selection}. The input of attention in \textit{HDR-NMT\_selection} has been modified as:    
\begin{equation}
Att(Q,K,V)=softmax( \frac{Q\cdot K_{HN}^T}{ \sqrt{d_k} } ) \cdot V_N
\end{equation}
\begin{equation}
K_{HN}=[K_H;K_N]
\end{equation}
Table~\ref{selection} illustrates the results of the HDR-NMT\_selection with a positive outcome. 

\begin{table}[h]\footnotesize
\centering
\setlength\tabcolsep{3pt}{
\begin{tabular}{lcccc}
\hline
\textbf{Model} & \textbf{EN $ \rightarrow $ RU} & \textbf{EN $ \rightarrow $ ZH} & \textbf{EN $ \rightarrow $ DE} & \textbf{EN $ \rightarrow $ FR}  \\ \hline
Fairseq & 23.9  & 25.5 & 26 & 38.3 \\ 
SR  & 25.0  & 26.2  & 26  & 38.5 \\
SR+WDR  & 25.4  & 26.1 & 26.1 & 38.6 \\ \hline
\end{tabular}}
\caption{The specific results of HDR\_NMT\_selection.}
\label{selection}
\end{table}

\subsection{Details of Datasets}
\label{detail}

The specific details of the involved datasets are depicted in Table~\ref{dataset}.

\begin{table}[H]\footnotesize
\centering
\begin{tabular}{lcccc}
\hline
\textbf{Language} & \textbf{\#Train} & \textbf{\#Dev} & \textbf{\#Test} & \textbf{\#Vocab} \\ \hline
EN $ \rightarrow $ RU  & 21M  & 3000 & 3001 & 25k/31k \\
EN $ \rightarrow $ ZH  & 8M   & 5982 & 2001 & 19k/49k \\ 
EN $ \rightarrow $ DE  & 3.5M & 36023 & 2169 & 30k/31k \\ 
EN $ \rightarrow $ FR  & 30M  & 23102 & 3003 & 31k/31k \\ \hline
\end{tabular}
\caption{The specific count of the training, development, and test sets. The values in \#Vocab show the size of the source and target vocabularies.}
\label{dataset}
\end{table}

\subsection{Total Number of Parameters}
\label{number}

Table~\ref{para} shows the total number of parameters for all models. 

\begin{table}[h]\footnotesize
\centering
\begin{tabular}{lccccc}
\hline
\textbf{Language} & \textbf{add} & \textbf{gate} & \textbf{cascade} & \textbf{selection} & \textbf{baseline} \\ \hline
EN $ \rightarrow $ RU  & 111k  & 98k  & 98k  & 100k & 73k \\
EN $ \rightarrow $ ZH  & 117k  & 104k & 104k & 106k & 79k \\ 
EN $ \rightarrow $ DE  & 113k  & 100k & 100k & 102k & 75k \\ 
EN $ \rightarrow $ FR  & 114k  & 101k & 101k & 103k & 76k \\ \hline
\end{tabular}
\caption{The specific total number of parameters in every HDR-NMT model. The ``add'' represents the HDR-NMT\_add, ``gate'' represents the HDR-NMT\_gate, ``cascade'' represents the HDR-NMT\_cascade and ``selection'' represents the HDR-NMT\_selection.}
\label{para}
\end{table}

\subsection{Effects of SR and WDR Learning on More Translation Tasks}
\label{effect}

Tables~\ref{en2zh}-\ref{en2de} demonstrate the effects of SR and WDR learning can be observed on other translation tasks (EN $ \rightarrow $ ZH and EN $ \rightarrow $ DE). 

\begin{table}[h]
\centering
\setlength\tabcolsep{3pt}{
\begin{tabular}{lccc}
\hline
\textbf{EN $ \rightarrow $ ZH} & \textbf{add} & \textbf{gate} & \textbf{cascade}  \\ \hline
SR  & 25.9  & 26  & 25.9   \\
SR+WDR  & 26.2  & 25.8 & 25.7  \\ 
FT-SR  & 26.1  & 26.3 & 26.5  \\ 
FT-SR+WDR  & 26.2  & 26.6 & 26.4 \\ \hline
\end{tabular}}
\caption{The specific results of EN $ \rightarrow $ ZH translation task.}
\label{en2zh}
\end{table}

\begin{table}[H]
\centering
\setlength\tabcolsep{3pt}{
\begin{tabular}{lcccc}
\hline
\textbf{EN $ \rightarrow $ DE} & \textbf{add} & \textbf{gate} & \textbf{cascade}  \\ \hline
SR  & 26.3  & 25.7  & 26   \\
SR+WDR  & 26.4  & 26.2 & 26.1 \\ 
FT-SR  & 26.6  & 26.3 & 26.4  \\ 
FT-SR+WDR  & 26.9  & 26.7 & 26.5  \\ \hline
\end{tabular}}
\caption{The specific results of EN $ \rightarrow $ DE translation task.}
\label{en2de}
\end{table}

\subsection{Additional NMT Encoder Does not Help}
\label{additional}
Table~\ref{dual} shows that simply adding an NMT-encoder does not produce a postive outcome.

\begin{table}[h]\footnotesize
\centering
\setlength\tabcolsep{3pt}{
\begin{tabular}{lc}
\hline
\textbf{Model} & \textbf{EN $ \rightarrow $ ZH}   \\ \hline
Fairseq  & 25.5    \\ 
NMT-encoder+NMT-encoder  & 25.0(-0.5)  \\
HDR-encoder+NMT-encoder & 26.2(+0.7)  \\  \hline
\end{tabular}}
\caption{The comparison of effect on different dual NMT-encoders.}
\label{dual}
\end{table}

\begin{table}[h]
\centering
\setlength\tabcolsep{3pt}{
\begin{tabular}{lc}
\hline
\textbf{Model} & \textbf{EN $ \rightarrow $ ZH}   \\ \hline
Fairseq  & 6.050    \\ 
HDR-NMT & 7.700  \\
Reference & 8.075  \\  \hline
\end{tabular}}
\caption{The human evaluation results on EN $ \rightarrow $ ZH translation task.}
\label{human}
\end{table}


\begin{figure*}[h]
\centering                                                          
\subfigure[baseline-encoder]{\label{baseline-encoder-logistics}\includegraphics[scale=0.6]{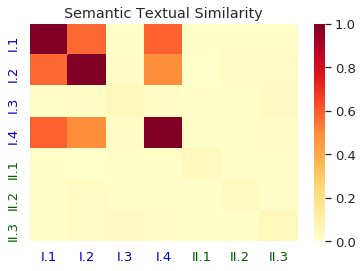}}               
\subfigure[HDR-encoder]{\label{NMT-encoder-logistics}\includegraphics[scale=0.6]{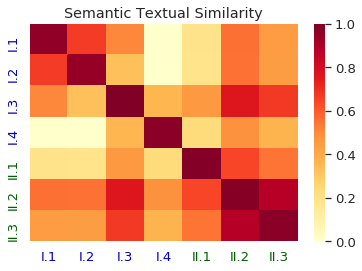}}
\caption{(a) The heatmap results of the baseline encoder. (b) The heatmap result of the HDR-encoder. \color{blue}I.1 \color{black}: The main logistics base will be established in Nyala. \color{blue}I.2 \color{black}: Deploy experienced in-country logistics staff, as required. \color{blue}I.3 \color{black}: Length of logistics chains. \color{blue}I.4 \color{black}: Traceability and logistics of operations. \color{green}II.1 \color{black}: Achieving this would require significant improvement of logistics capabilities in the region. \color{green}II.2 \color{black}: Distribution services are central in the logistics chain and for trade in goods and other services. \color{green}II.3 \color{black}: Transport, logistics and global value chains.}      
\label{logistics}                                   
\end{figure*}

\subsection{More Visualization and Translation Examples}
\label{vis}
Figure~\ref{logistics} and Table~\ref{example2} capture more visualization and translation examples demonstrating the effects of the proposed method.

\begin{table*}[t]
\begin{tabular}{p{95pt}p{80pt}p{80pt}p{80pt}p{80pt}}
\hline 
\specialrule{0em}{1pt}{1pt}
\textbf{Source}  & \textbf{Baseline Output}  & \textbf{HDR-NMT$_{SR}$ Output}   & \textbf{HDR-NMT$_{SR+WDR}$ Output}  & \textbf{Ground Truth}  \\ \hline 
\specialrule{0em}{1pt}{1pt}
\color{red}\textbf{Sims} \color{black} was 20 years old. & \color{red}\textbf{模拟} \color{black}已经20岁。    & \color{red}\textbf{模拟人} \color{black}年仅20岁。 & \color{red}\textbf{西姆斯} \color{black}20岁。   & \color{red}\textbf{Sims} \color{black} 享年 20 岁。  \\ \hline 
\specialrule{0em}{1pt}{1pt}
We're \color{red}\textbf{doing great with} \color{black} North Korea, he said.  & 他说,“ 我们在北朝鲜问题上\color{red}\textbf{做得很好} \color{black}。”   & 他说,“我们在北朝鲜\color{red}\textbf{做了大量工作} \color{black}”。 & 他说,“ 我们与北朝鲜\color{red}\textbf{合作得不错} \color{black}。 ”    & 他说道，“我们跟朝鲜\color{red}\textbf{相处得很好} \color{black}。” \\ \hline 
\specialrule{0em}{1pt}{1pt}
He is one of a number of famous \color{red}\textbf{rappers} \color{black} to change their name.     & 他是许多更改姓氏的著名\color{red}\textbf{拉客} \color{black}之一。       & 他是一些著名的改名者之一。      & 他是一些改变姓氏的著名\color{red}\textbf{歌手} \color{black}之一。       & 很多著名\color{red}\textbf{说唱歌} \color{black}手都改过名字。  \\ \hline 
\specialrule{0em}{1pt}{1pt}
They said, 'don't go out there\color{red}\textbf{with that hat on} \color{black}.'   & 他们说,'不要\color{red}\textbf{随身而出} \color{black}。   & 他们说,'不要\color{red}\textbf{带着这种头衔离开那里} \color{black}。 & 他们说,'不要\color{red}\textbf{戴上这个帽子出去} \color{black}'。   & 他们说，‘别\color{red}\textbf{带着那顶帽子出去} \color{black}'。\\ \hline 
\specialrule{0em}{1pt}{1pt}
They want \color{red}\textbf{to be famous to say my name} \color{black}, but it is part of the job. & 他们想要\color{red}\textbf{说出自己的名字} \color{black},但这是工作的一部分。 & 他们\color{red}\textbf{想要有名声} \color{black},但这是工作的一部分  & 他们\color{red}\textbf{想用我的名字来闻名} \color{black},但这是工作的一部分。 & 他们想\color{red}\textbf{利用我出名} \color{black}，但这也是他们工作的一部分。       \\ \hline 
\specialrule{0em}{1pt}{1pt}
\color{red}\textbf{Red Tide} \color{black} has also been observed inPasco County.     & 在帕斯科州也发现了\color{red}\textbf{红色赛德} \color{black}。         & 帕斯科州也观察到\color{red}\textbf{红铁德} \color{black}        & 在帕斯科州也观察到\color{red}\textbf{红潮} \color{black}。           & 帕斯科县 ( Pasco County) 也出现了\color{red}\textbf{赤潮} \color{black}。\\ \hline  

\end{tabular}
\caption{The comparable examples extracted from the outcomes for the newstest19-EN $ \rightarrow $ ZH.}
\label{example2}
\end{table*}

\end{CJK}
\end{document}